\documentclass[runningheads]{llncs}
\usepackage[T1]{fontenc}
\usepackage{graphicx}
\usepackage{pifont} 
\usepackage{xcolor}
\usepackage{xcolor}
\usepackage{amsmath, amssymb}
\usepackage{graphicx}
\usepackage{hyperref} 
\usepackage{xcolor}   
\usepackage{float}

\usepackage{orcidlink}    
\usepackage{amssymb}
\usepackage{marvosym}
\usepackage[normalem]{ulem}
\useunder{\uline}{\ul}{}
\hypersetup{
    colorlinks=true,        
    linkcolor=blue,         
    filecolor=magenta,      
    urlcolor=blue,          
    citecolor=green,        
    pdfborder={0 0 0}       
}

\begin{document}
\title{Graph-Based Uncertainty Modeling and Multimodal Fusion for Salient Object Detection}
\titlerunning{Uncertainty Modeling and Multimodal Fusion for SOD}

\author{
Yuqi Xiong \inst{1} \orcidlink{0009-0001-4204-9385} \and 
Wuzhen Shi \inst{1} \orcidlink{0000-0002-6819-0125} \and
Yang Wen \inst{1} \textsuperscript{$($\Letter$)$} \orcidlink{0000-0001-6303-8178} \and Ruhan Liu \inst{2} \orcidlink{0000-0002-0281-8039}
}

\authorrunning{Y. Xiong et al.}

\institute{
Guangdong Key Laboratory of Intelligent Information Processing,\\
College of Electronics and Information Engineering, Shenzhen University,\\ Shenzhen, China \\
\email{2022090048@email.szu.edu.cn, wzhshi@szu.edu.cn, wen\_yang@szu.edu.cn}
\and
Furong Laboratory, Central South University, Changsha, China \\
\email{223101@csu.edu.cn}
}

\maketitle

\begin{abstract}
In view of the problems that existing salient object detection (SOD) methods are prone to losing details, blurring edges, and insufficient fusion of single-modal information in complex scenes, this paper proposes a dynamic uncertainty propagation and multimodal collaborative reasoning network (DUP-MCRNet). Firstly, a dynamic uncertainty graph convolution module (DUGC) is designed to propagate uncertainty between layers through a sparse graph constructed based on spatial semantic distance, and combined with channel adaptive interaction, it effectively improves the detection accuracy of small structures and edge regions. Secondly, a multimodal collaborative fusion strategy (MCF) is proposed, which uses learnable modality gating weights to weightedly fuse the attention maps of RGB, depth, and edge features. It can dynamically adjust the importance of each modality according to different scenes, effectively suppress redundant or interfering information, and strengthen the semantic complementarity and consistency between cross-modalities, thereby improving the ability to identify salient regions under occlusion, weak texture or background interference. Finally, the detection performance at the pixel level and region level is optimized through multi-scale BCE and IoU loss, cross-scale consistency constraints, and uncertainty-guided supervision mechanisms. Extensive experiments show that DUP-MCRNet outperforms various SOD methods on most common benchmark datasets, especially in terms of edge clarity and robustness to complex backgrounds. Our code is publicly available at \href{https://github.com/YukiBear426/DUP-MCRNet}{https://github.com/YukiBear426/DUP-MCRNet}.


\end{abstract}

\section{Introduction}

Salient Object Detection (SOD) aims to mimic the human visual system by automatically locating the most attention-grabbing regions in an image. It has broad applications in object recognition \cite{Zhou_2021}, semantic segmentation \cite{sun2020miningcrossimagesemanticsweakly}, and visual fixations \cite{wang2021salientobjectdetectiondeep}. Early SOD approaches relied on hand-crafted low-level cues—color contrast, texture, and edges—which work in simple scenes but falter in complex backgrounds due to the absence of high-level semantics.

With the rise of deep learning, CNN-based SOD methods have flourished. DSS \cite{Hou_2019} introduced deep supervision and short connections within the HED backbone to fuse multi-scale features; Amulet \cite{8237293} proposed a pyramid fusion strategy to merge spatial details and semantics across levels; PiCANet \cite{9076883} added an explicit attention module to adaptively model long- and short-range pixel dependencies, enriching local perception. However, fixed-stride down- and up-sampling in these architectures creates an inherent trade-off between preserving high-frequency details and capturing semantic context.

Recently, inspired by Transformers’ success in NLP, self-attention has propelled SOD forward. VST \cite{liu2021visualsaliencytransformer} was the first to employ a Transformer backbone, capturing long-range dependencies and bolstering global consistency; SRFormer \cite{10377606} combined Permuted Self-Attention with deep convolutions to mitigate semantic shifts across feature levels; TransformerSOD \cite{10697198} designed efficient local–global adaptive blocks to balance accuracy and speed. Yet most still use a uniform receptive field, failing to adjust feature complexity per region, which causes blurring or missed detections on fine structures like edges or small objects.

Moreover, most SOD works remain unimodal, overlooking the complementary power of multimodal data in challenging scenes. S2MA \cite{9156287} and BBSNet \cite{Zhai_2021} show that depth cues compensate for RGB’s weaknesses in low-contrast or occluded settings; SAD \cite{cen2023sadsegmentrgbd} explores cross-modal reasoning via semantic boundaries and depth guidance, improving robustness. Still, multimodal methods typically process each modality in isolation and lack an effective consistency-driven fusion scheme.

To address these gaps, we propose DUP-MCRNet: a Dynamic Uncertainty Propagation and Multimodal Collaborative Reasoning framework. First, we introduce Dynamic Uncertainty Graph Convolution, which builds a sparse graph based on spatial locality and semantic similarity to propagate uncertainty dynamically across positions, enhancing edge and small-object saliency. Second, we propose a Multimodal Collaborative Fusion strategy that employs learnable modality weights to adaptively fuse attention maps from RGB, depth, and edge features.Our contributions are as follows:

\begin{enumerate}
    \item \textbf{Dynamic Uncertainty Graph Convolution:}  
    Build a sparse graph on spatial–semantic distances to propagate uncertainty across levels, plus channel-wise dynamic fusion to preserve details and semantics.

    \item \textbf{Multimodal Collaborative Fusion:}  
    Use learnable weights to adaptively combine RGB, depth, and edge features, enhancing cross-modal complementarity and robustness in complex scenes.

    \item \textbf{Empirical Validation:}  
    Show that DUP-MCRNet consistently outperforms state-of-the-art SOD methods in accuracy and robustness on standard benchmarks.
\end{enumerate}

\section{Background}

\subsection{Multi-scale Feature Fusion}
Early salient object detection methods are based on traditional image processing methods such as GC \cite{6871397} to extract single-scale features or use convolutional neural networks such as DSS \cite{Hou_2019} to directly predict saliency through shallow or mid-level features. Although these methods can extract local textures, they lack understanding of global structures, resulting in blurred boundaries and poor internal consistency of detection results. In order to overcome the limitation of single scale, Amulet \cite{8237293} and RA \cite{chen2019reverseattentionsalientobject} proposed a saliency detection framework based on multi-scale feature aggregation, which fuses features of different resolutions to take into account both fine-grained details and high-level semantic information. However, most of these methods use static splicing or simple layer-by-layer addition, and fail to dynamically adjust the fusion weights according to the uncertainty of spatial position, resulting in insufficient perception of complex boundaries and small objects. PA2Net \cite{10.1007/978-981-96-2064-7_14} combines pyramid attention and recursive aggregation strategies to capture salient object information at multiple scales, partially alleviating the problem of spatial information loss during the fusion process. AFNet \cite{8954094} proposes a feedback mechanism, introduces salient region attention feedback, and guides low-level features to learn more accurate boundary information. However, these methods still use global unified weighting and cannot achieve local adaptive fusion of fuzzy boundary areas, detail-rich areas, and smooth areas. Recently, EGNet \cite{9008371} uses a boundary guidance module to enhance the boundary perception of salient objects in the feature encoding stage, while BASNet \cite{8953756} models saliency detection as a residual learning task to gradually refine the details of salient areas. Although these methods have made progress in boundary preservation, they still suffer from boundary detail loss in complex scenarios due to the lack of modeling of feature uncertainty relationships. To this end, this paper proposes a dynamic uncertainty graph convolution and channel adaptive interaction module, which models spatial uncertainty associations through sparse graphs and adaptively guides feature propagation, effectively alleviating the limitations of traditional static fusion methods in fine-grained structure modeling.

\subsection{Multimodal Collaborative Reasoning}
In order to further improve detection accuracy, researchers have begun to introduce multimodal information to assist saliency detection in recent years. DF \cite{Qu_2017} and PCF \cite{8578420} pioneered the exploration of RGB-Depth fusion, using the geometric structure of depth maps to assist saliency reasoning. However, their fusion strategies are mostly simple splicing or early fusion, which fails to fully explore the complementary characteristics between multiple modalities.

CoNet \cite{10.1007/978-3-030-58523-5_4} introduced a collaborative attention mechanism, which encodes RGB and Depth features separately and then performs high-order correlation modeling, effectively improving cross-modal reasoning capabilities. DMRA \cite{9010728} proposed a multi-scale recursive attention mechanism, which uses depth information to dynamically adjust the attention area of RGB features. In terms of receptive field modeling, traditional methods such as F³Net \cite{wei2019f3netfusionfeedbackfocus} and MINet \cite{pang2020multiscaleinteractivenetworksalient} mainly rely on multi-scale feature interactions at a fixed resolution, which makes it difficult to take into account both fine-grained areas and large-scale context modeling. Swin Transformer \cite{9710580} and PVT \cite{wang2021pyramidvisiontransformerversatile} introduced local perception capabilities through local window attention and pyramid structure. Inspired by these, this paper proposes a multimodal collaborative fusion, which dynamically adjusts the importance of RGB, depth, and edge features through a learnable modal gating mechanism, effectively improving the saliency detection effect in complex scenes.

\section{Methodology}
\begin{figure}
    \centering
    \includegraphics[width=1\linewidth]{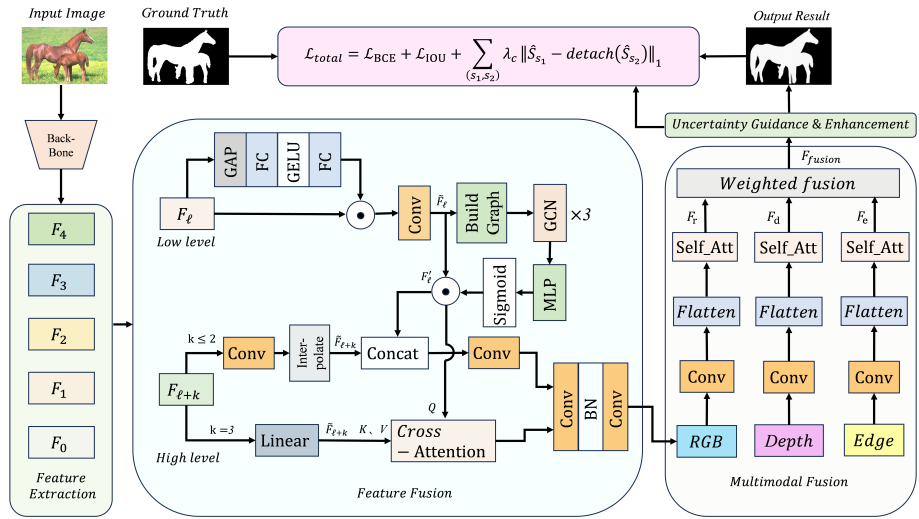}
    \caption{The overview architecture of our proposed model DUP-MCRNet. Black arrows indicate the data flow. Circle with a dot inside denotes element-wise multiplication.}
    \label{fig:architecture}
\end{figure}

Figure~\ref{fig:architecture} shows the overview architecture of our model. First, the input image is passed through the CNN or Transformer based backbone to extract features of different scales. Then, the features of different levels are passed in turn to the dynamic uncertainty graph convolution and channel adaptive interaction module we proposed to interact with information of different scales. Then, the multimodal collaborative fusion part will use the RGB, Depth and Edge information in the feature map, and perform weighted fusion after self-attention to enhance the model's utilization of multimodal information in the image. Finally, the final output is obtained through an uncertainty enhancement module.

\subsection{Dynamic Uncertainty Graph Interaction}

In this module, channel attention is first introduced to the input low-level feature tensor $F_\ell\in\mathbb{R}^{B\times C\times H\times W}$, where $B$ is the batch size, $C$ is the number of channels, representing the feature dimension of each pixel position, $H$ is the height of the feature map, and $W$ is the width of the feature map. It is then mapped to a unified embedding dimension $d$ through a \(1 \times 1\) convolution to form $\tilde{F}_\ell\in\mathbb{R}^{B\times d\times H\times W}$. Subsequently, $\tilde{F}$ is flattened into a feature sequence of length \( N = H \times W \) according to the spatial position and the order is exchanged with the dimensions to obtain $X\in\mathbb{R}^{B\times N\times d}$. Next, a sparse graph is constructed on these spatial nodes for each batch. Specifically, the comprehensive distance $D_{\mathrm{comb}}(i,j)$ is first calculated by a weighted combination of the spatial coordinate distance and the feature similarity, as shown in Formula~\eqref{eq:1}:

\begin{equation}
D_{\mathrm{comb}}(i,j)=\alpha D_{\mathrm{spatial}}(i,j)+(1-\alpha)D_{\mathrm{feature}}(i,j)
\label{eq:1}
\end{equation}

where $\alpha\in(0,1)$ is the weight, $i$,$j$ represents two different spatial positions in the feature map, $D_{\mathrm{spatial}}(i,j)$ is the spatial coordinate distance, and $D_{\mathrm{feature}}(i,j)$ is the feature similarity. The calculation method is shown in Formula~\eqref{eq:2}:

\begin{equation}
D_{\mathrm{spatial}}(i,j)=\left\|P_i-P_j\right\|_2\quad D_{\mathrm{feature}}(i,j)=1-\frac{X_i\cdot X_j}{\|X_i\|\|X_j\|}
\label{eq:2}
\end{equation}

According to $D_{\mathrm{comb}}(i,j)$, the smallest $K$ neighbors are selected for each row to generate the adjacency matrix, as shown in Formula~\eqref{eq:3}:

\begin{equation}
A_{ij} =
\begin{cases}
1, & D_{\mathrm{comb}}(i,j)\in\mathrm{TopK}(D_{\mathrm{comb}}(i,:)) \\
0, & \text{otherwise}
\end{cases}
\label{eq:3}
\end{equation}

To ensure self-loop information, we set $\tilde{A}=A+I$ and normalize it with the degree matrix $\widetilde{D}_{ii}=\sum_j\tilde{A}_{ij}$. After that, the iterative propagation of uncertainty along the graph structure is explicitly completed through three layers of graph convolution, as shown in Formula~\eqref{eq:4}:

\begin{equation}
X^{(t+1)}=X^{(t)}+\mathrm{ReLU}\left(\widetilde{D}^{-\frac{1}{2}}\widetilde{A}\widetilde{D}^{-\frac{1}{2}}X^{(t)}W_g^{(t)}\right),t=0,1,2
\label{eq:4}
\end{equation}

where $X^{(t)}$ is the feature representation of each node in the $t$-th graph convolution iteration, and the result is finally reshaped back to the spatial shape $X^{(3)}\in\mathbb{R}^{B\times d\times H\times W}$. In order to highlight the impact of the uncertain region on the subsequent saliency representation, we generate a pixel-level uncertainty weight map through a lightweight MLP and multiply it with $\tilde{F}_{\ell}$ to obtain $F_\ell^{\prime}\in\mathbb{R}^{B\times d\times H\times W}$, as shown in Formula~\eqref{eq:5}:

\begin{equation}
F_\ell^{\prime}=\tilde{F}_\ell\odot\left(1+\sigma\left(\mathrm{MLP}(X^{(3)})\right)\right)
\label{eq:5}
\end{equation}

After completing graph convolution and uncertainty enhancement, the module performs channel adaptive compression and spatial alignment on the high-level features $F_{\ell+k}$. For adjacent levels $(k\leq2)$, first reduce its channel to dimension-$d$ by \(1 \times 1\) convolution, and then bilinearly interpolate to (H,W), and obtain $\tilde{F}_{\ell+k}\in\mathbb{R}^{B\times d\times H\times W}$. $\tilde{F}_{\ell+k}$ and $F_\ell^{\prime}$ are concatenated in the channel dimension, and then two \(3 \times 3\) convolutions and Relu activation are performed to obtain the fusion feature $F_{\mathrm{fuse}}$; for cross-layer layers $(k=3)$, $F_{\ell+k}$ is channel-projected to obtain $\tilde{F}_{\ell+k}\in\mathbb{R}^{B\times d\times H\times W}$, which is mapped to the $Key$, $Value$ and $Query$ spaces with the low-level feature $\tilde{F}_{\ell}$ respectively. Finally, the cross-layer semantic fusion is completed using the cross-attention mechanism, as shown in Formula~\eqref{eq:6}:

\begin{equation}
Attention=\mathrm{Softmax}\left(\frac{(F_\ell^{\prime}W_q)(\tilde{F}_{\ell+k}W_k)^{\mathsf{T}}}{\sqrt{d}}\right)\left(\tilde{F}_{\ell+k}W_v\right)
\label{eq:6}
\end{equation}

$W_q,W_k,W_v$ are learnable parameter matrices. This method uses a cross attention mechanism \cite{xiong2025nonstationarytimeseriesforecasting} to fuse features from different levels, using high-level semantics as $Key$ and $Value$ and low-level spatial features as $Query$. This asymmetric attention mechanism exploits the complementarity of scales: global semantics guides the selective extraction of local detail information. Compared with simple concatenation or summation, the cross attention mechanism dynamically weights spatial positions, thereby more accurately fusing semantics and details, enhancing feature discrimination and context awareness.

The obtained attention output is restored to its original shape through linear mapping and then added back to $F_\ell^{\prime}$ to complete the interaction, obtaining $F_{\mathrm{fuse}}$. Finally, all cross-layer interaction results are aggregated, and the obtained $F_{\mathrm{fuse}}^{(k)}$ is passed through a set of bottleneck convolutions to obtain the final output $F_{\mathrm{out}}\in\mathbb{R}^{B\times C\times H\times W}$, as shown in Formula~\eqref{eq:7}:

\begin{equation}
F_{\mathrm{out}}=\mathrm{Conv}_{1\times1}\left(\mathrm{RELU}\left(\mathrm{Conv}_{3\times3}\left(F_{\mathrm{fuse}}^{(k)}\right)\right)\right)
\label{eq:7}
\end{equation}

Through the above design, the model not only explicitly models the uncertainty transmission between different levels in the graph structure, but also tailors the channel fusion strategy for each layer feature, thereby achieving significant improvements in maintaining high-frequency details and cross-layer semantic consistency.

\subsection{Multimodal Collaborative Fusion}

First, for the cross-modal information fusion part, we perform channel mapping and attention encoding on the three inputs of RGB, Depth and Edge respectively. Assuming that the three modal inputs are $F_{\mathrm{out}}{}^{(m)}\in\mathbb{R}^{B\times C\times H\times W}$, where $m=1$ represents RGB, $m=2$ represents Depth, and $m=3$ represents Edge, multimodal collaborative fusion model maps the $m$-th modality to a unified embedding dimension through \(1 \times 1\) convolution, as shown in Formula~\eqref{eq:8}:

\begin{equation}
F^{(m)}=\mathrm{Conv}_{1\times1}\left(F_{\mathrm{out}}^{(m)}\right)\in\mathbb{R}^{B\times d\times H\times W}
\label{eq:8}
\end{equation}

Then it is flattened according to the spatial dimension to $F_{\mathrm{flat}}^{(m)}\in\mathbb{R}^{B\times d\times N}$, where \( N = H \times W \), and then the order of the channel and spatial dimensions is interchanged to obtain the sequence representation $F_{\mathrm{seq}}^{(m)}\in\mathbb{R}^{B\times N\times d}$. Next, the self-attention mechanism is performed on each modal sequence to obtain the attention output $\hat{F}_{\mathrm{seq}}^{(m)}\in\mathbb{R}^{B\times N\times d}$. Finally, this sequence is reshaped back to the original spatial size ${\hat{F}}^{(m)}\in\mathbb{R}^{B\times d\times H\times W}$ to complete the global spatial attention encoding of the $m$-th modality. At the same time, in order to allow the network to automatically learn the importance of each modality, we introduce a learnable modality weight parameter $\theta\in\mathbb{R}^3$ to obtain the weight $w_{m}$, and weightedly fuse the attention graphs of each modality, as shown in Formula~\eqref{eq:9}:

\begin{equation}
w_m=\frac{\exp{(\theta_m)}}{\sum_{k=1}^3\exp{(\theta_k)}}\quad F_{\mathrm{fus}}=\sum_{m=1}^3w_m\widehat{F}^{(m)}
\label{eq:9}
\end{equation}

This fusion method not only retains the complementarity of multi-source information, but also can adaptively adjust the modality weights during training, thereby obtaining more accurate salient area positioning in scenarios such as blurred boundaries and drastic depth changes.

\subsection{Loss Function}

To effectively guide the model in learning accurate localization and structural detail restoration of salient objects, we design a comprehensive loss function system that integrates hierarchical supervision and fine-grained regularization. The objective is to jointly optimize local pixel-wise accuracy, regional structural consistency, and boundary refinement, thereby enhancing overall prediction quality. The proposed loss function system consists of the following components:

First, to enhance the model’s ability to capture multi-level semantics and structural cues, we generate saliency outputs $\{S_{1/16}, S_{1/8}, S_{1/4}\}$ from different decoder stages, corresponding to progressively higher resolutions. These outputs supervise the model in localizing salient regions at multiple scales. In parallel, we design an uncertainty-aware mask prediction branch that produces mask outputs $\{M_{1/4}, M_{1/2}, M_{1/1}\}$, which encode spatial uncertainty and assist in boundary refinement.

For these multi-scale saliency and mask outputs, we employ scale-wise supervision using binary cross-entropy (BCE) and intersection-over-union (IoU) losses. The BCE loss is defined as shown in Formula~\eqref{eq:loss_bce}:

\begin{equation}
\mathcal{L}_{\mathrm{BCE}}(\hat{S},Y)=-\frac{1}{N}\sum_{i=1}^{N}\left(Y_{i}\log(\hat{S}_{i})+(1-Y_{i})\log(1-\hat{S}_{i})\right)
\label{eq:loss_bce}
\end{equation}

where $N$ is the total number of pixels, $\hat{S}$ is the predicted output, and $Y$ is the ground-truth label. To further improve the region-level coherence of the predicted mask, the IoU loss is introduced, as defined in Formula~\eqref{eq:loss_iou}:

\begin{equation}
\mathcal{L}_{\mathrm{IoU}}(\hat{S},Y)=1-\frac{\sum_{i=1}^{N}\hat{S}_{i}Y_{i}}{\sum_{i=1}^{N}(\hat{S}_{i}+Y_{i}-\hat{S}_{i}Y_{i})}
\label{eq:loss_iou}
\end{equation}

Based on these two basic components, the overall saliency loss $\mathcal{L}_{\mathrm{sal}}$ is defined in Formula~\eqref{eq:loss_sal}:

\begin{equation}
\begin{aligned}
\mathcal{L}_{\mathrm{sal}} &= \sum_{s\in\{1/16,1/8,1/4\}} \left( \mathcal{L}_{\mathrm{BCE}}(S_{s},Y) + \mathcal{L}_{\mathrm{IoU}}(S_{s},Y) \right) \\
&+ \sum_{m\in\{1/4,1/2,1/1\}} \left( \mathcal{L}_{\mathrm{BCE}}(M_{m},Y) + \mathcal{L}_{\mathrm{IoU}}(M_{m},Y) \right)
\end{aligned}
\label{eq:loss_sal}
\end{equation}

Here, $S_{s}$ denotes saliency maps at different scales, and $M_{m}$ denotes the corresponding uncertainty-aware mask outputs. To mitigate potential prediction bias between outputs of different scales, we introduce a cross-scale consistency constraint. For adjacent scale pairs $(s_1, s_2)$, the consistency loss is defined in Formula~\eqref{eq:loss_consistency}:

\begin{equation}
\mathcal{L}_{\mathrm{consistency}} = \sum_{(s_1,s_2)} \lambda_c \left\| \hat{S}_{s_1} - \mathrm{detach}(\hat{S}_{s_2}) \right\|_1
\label{eq:loss_consistency}
\end{equation}

where $\lambda_c$ is a weighting coefficient, and $\mathrm{detach}(\cdot)$ prevents gradient propagation to higher-level predictions, ensuring that lower-level features are learned independently. Combining the above losses, our total loss function is expressed as Formula~\eqref{eq:total_loss}:

\begin{equation}
\mathcal{L}_{\mathrm{total}}=\mathcal{L}_{\mathrm{sal}}+\mathcal{L}_{\mathrm{consistency}}
\label{eq:total_loss}
\end{equation}

\section{Experiment}

\subsection{Datasets Description}
We evaluate our model on six widely used benchmark datasets. SOD \cite{b40} contains 300 images from BSD, which contains multiple low-contrast salient objects, often overlapping with image boundaries, and is highly challenging. ECSSD \cite{b42} contains 1000 complex scene images, most of which contain a single salient object, with diverse foreground and background patterns, from BSD, VOC2012, and the Internet. PASCAL-S \cite{b44} is based on 850 images from PASCAL VOC 2010, and adds eye gaze points and salient segmentation annotations on top of the original annotations. DUT-OMRON \cite{b46} contains 5168 images, covering complex backgrounds and multiple salient objects, and provides accurate pixel-level annotations. HKU-IS \cite{b47} contains 4447 images with salient object annotations, and the images meet at least one of the following conditions: there are multiple unconnected objects, objects touch edges, or the color contrast is less than 0.7. DUTS \cite{b48} is the largest saliency detection dataset currently, containing 10,553 training images and 5,019 test images, selected from the ImageNet and SUN datasets respectively. All annotations are manually completed by 50 participants.

\subsection{Experimental Setup}

We use 10,553 images from the DUTS dataset for training, and resize all images to $384 \times 384$. The batch size is set to 8, and the model is trained for 100 epochs. The Adam optimizer \cite{b50} is employed with an initial learning rate of $1\times10^{-4}$, and a learning rate decay strategy is applied to facilitate convergence. All experiments are conducted on an NVIDIA RTX 4090 GPU with 24 GB memory.

To comprehensively evaluate the performance of the proposed method, we adopt four mainstream evaluation metrics: mean absolute error (MAE), mean intersection-over-union (mSIOU), structural similarity measure (S-measure), and weighted F-measure (Weighted-$F_\beta$). The specific definitions of these metrics are provided below:

\subsubsection{Mean Absolute Error (MAE):}  
MAE measures the average absolute pixel-wise difference between the predicted saliency map $P$ and the ground truth label $G$, and is defined as shown in Formula~\eqref{eq:mae}:

\begin{equation}
\mathrm{MAE} = \frac{1}{H \times W} \sum_{x=1}^{H} \sum_{y=1}^{W} |P(x, y) - G(x, y)|
\label{eq:mae}
\end{equation}

where $H$ and $W$ denote the height and width of the image, respectively. $P(x, y)$ and $G(x, y)$ represent the predicted and ground truth values at pixel $(x, y)$. A smaller MAE indicates better prediction accuracy.

\subsubsection{Mean Structural IoU (mSIOU):}  
mSIOU evaluates the structural overlap between predicted regions and ground truth, reflecting the consistency at a structural level. It is computed as shown in Formula~\eqref{eq:msiou}:

\begin{equation}
\mathrm{mSIOU} = \frac{1}{N} \sum_{i=1}^{N} \frac{P_i \cap G_i}{P_i \cup G_i}
\label{eq:msiou}
\end{equation}

In this equation, $P_i$ and $G_i$ denote the predicted and ground truth binary regions for the $i$-th instance, and $N$ is the total number of pixels or segments. Higher mSIOU values indicate better structural alignment.

\subsubsection{Structural Similarity Measure (S-measure):}  
S-measure combines object-level and region-level structural similarities into a unified metric. It is defined as shown in Formula~\eqref{eq:smeasure}:

\begin{equation}
S_m = \alpha S_o + (1 - \alpha) S_r
\label{eq:smeasure}
\end{equation}

where $S_o$ is the object-level structural similarity, $S_r$ is the region-level similarity, and $\alpha$ is a balancing coefficient. A higher $S_m$ value implies that both global and local structures are well preserved.

\subsubsection{Weighted F-measure (Weighted-$F_\beta$):}  
This metric extends the traditional F-measure by incorporating pixel-wise weights in the computation of precision and recall, which allows better handling of imbalanced and spatially variant errors. It is defined in Formula~\eqref{eq:weightedf}:

\begin{equation}
F_\beta^w = \frac{(1 + \beta^2) \cdot \mathrm{Precision}_w \cdot \mathrm{Recall}_w}{\beta^2 \cdot \mathrm{Precision}_w + \mathrm{Recall}_w}
\label{eq:weightedf}
\end{equation}

Here, $\beta^2$ adjusts the relative importance of precision and recall. Weighted-$F_\beta$ better captures the perceptual quality of predictions by considering both spatial and neighborhood error distributions, especially for salient objects with blurred or incomplete boundaries.

\subsection{Experimental Results}

Table~\ref{tab:quantitative_results} shows the quantitative comparison results on five benchmark datasets. We compare our method against five state-of-the-art models: CPD \cite{b51}, PoolNet \cite{b52}, LDF \cite{b53}, MINet \cite{pang2020multiscaleinteractivenetworksalient}, and UGRAN \cite{b54}. The four indicators of \textit{MAE}, \textit{mSIOU}, $S_m$, and $F_\beta^w$ are used to evaluate the performance of six methods. The results show that our method outperforms previous advanced methods in the indicators of most datasets.

\renewcommand{\arraystretch}{1.6}

\begin{table}[htbp]
\caption{Comparison results of different methods on five benchmark datasets. Bold indicates the best result and underline indicates the second best result.}
\centering
\resizebox{\textwidth}{!}{%
\begin{tabular}{c|cccc|cccc|cccc|cccc|cccc}
\hline
\textbf{Dataset} & \multicolumn{4}{c|}{\textbf{DUT-O}} & \multicolumn{4}{c|}{\textbf{DUTS}} & \multicolumn{4}{c|}{\textbf{ECSSD}} & \multicolumn{4}{c|}{\textbf{HKU-IS}} & \multicolumn{4}{c}{\textbf{PACSAL-S}} \\ \hline
\textbf{Metrics} & \textit{MAE}~\textcolor{red}{$\downarrow$} & \textit{mS}~\textcolor{green}{$\uparrow$} & $S_m$~\textcolor{green}{$\uparrow$} & $F_\beta^w$~\textcolor{green}{$\uparrow$} & \textit{MAE}~\textcolor{red}{$\downarrow$} & \textit{mS}~\textcolor{green}{$\uparrow$} & $S_m$~\textcolor{green}{$\uparrow$} & $F_\beta^w$~\textcolor{green}{$\uparrow$} & \textit{MAE}~\textcolor{red}{$\downarrow$} & \textit{mS}~\textcolor{green}{$\uparrow$} & $S_m$~\textcolor{green}{$\uparrow$} & $F_\beta^w$~\textcolor{green}{$\uparrow$} & \textit{MAE}~\textcolor{red}{$\downarrow$} & \textit{mS}~\textcolor{green}{$\uparrow$} & $S_m$~\textcolor{green}{$\uparrow$} & $F_\beta^w$~\textcolor{green}{$\uparrow$} & \textit{MAE}~\textcolor{red}{$\downarrow$} & \textit{mS}~\textcolor{green}{$\uparrow$} & $S_m$~\textcolor{green}{$\uparrow$} & $F_\beta^w$~\textcolor{green}{$\uparrow$} \\ \hline
CPD & 0.057 & 0.741 & 0.818 & 0.715 & 0.043 & 0.795 & 0.867 & 0.800 & 0.040 & 0.839 & 0.910 & 0.895 & 0.033 & 0.831 & 0.904 & 0.879 & 0.072 & 0.748 & 0.845 & 0.796 \\
PoolNet & \underline{0.056} & 0.754 & 0.836 & 0.729 & 0.040 & 0.807 & 0.883 & 0.807 & 0.039 & 0.851 & 0.921 & 0.896 & 0.032 & 0.850 & 0.917 & 0.883 & 0.075 & 0.766 & 0.849 & 0.723 \\
LDF & \textbf{0.052} & \underline{0.772} & \underline{0.839} & \underline{0.752} & \textbf{0.034} & 0.828 & 0.892 & 0.845 & \underline{0.034} & \underline{0.861} & \underline{0.924} & \underline{0.915} & \underline{0.028} & \underline{0.857} & 0.919 & 0.904 & \textbf{0.051} & \textbf{0.801} & \textbf{0.882} & \underline{0.847} \\
MINet & 0.057 & 0.753 & 0.822 & 0.718 & 0.039 & 0.804 & 0.875 & 0.813 & 0.036 & 0.857 & 0.919 & 0.905 & 0.031 & 0.853 & 0.912 & 0.889 & 0.064 & 0.763 & 0.854 & 0.808 \\
UGRAN & 0.058 & 0.768 & 0.830 & 0.733 & \textbf{0.034} & \textbf{0.856} & \textbf{0.924} & \textbf{0.911} & \underline{0.034} & 0.856 & \underline{0.924} & 0.911 & \underline{0.028} & 0.856 & \underline{0.922} & \underline{0.905} & \underline{0.059} & 0.773 & 0.867 & 0.826 \\
\hline
Ours & 0.066 & \textbf{0.788} & \textbf{0.840} & \textbf{0.755} & \underline{0.036} & \underline{0.845} & \underline{0.902} & \underline{0.858} & \textbf{0.030} & \textbf{0.881} & \textbf{0.935} & \textbf{0.922} & \textbf{0.027} & \textbf{0.876} & \textbf{0.929} & \textbf{0.913} & 0.061 & \underline{0.786} & \underline{0.869} & \textbf{0.930} \\ \hline
\end{tabular}%
}
\label{tab:quantitative_results}
\end{table}

\begin{figure}
    \centering
    \includegraphics[width=1\linewidth]{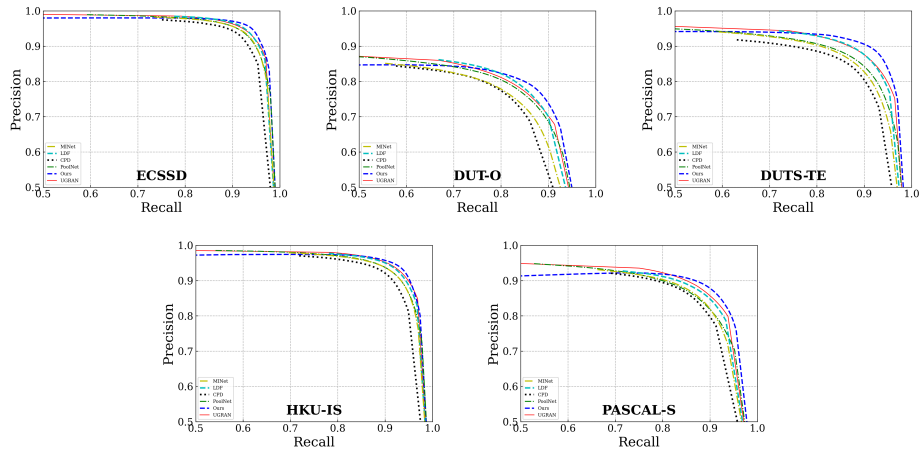}
    \caption{Precision-recall curves of different methods on the salient object detection task. The blue dashed line represents our model, and the other lines represent the models we compare against.}
    \label{fig:enter-label3}
\end{figure}

In Figure~\ref{fig:enter-label3}, we plot the precision-recall curves of each method to comprehensively evaluate the detection performance of the model at different thresholds. From the results, it can be seen that our method achieves a high balance between precision and recall, which verifies the superior performance and good robustness of the proposed model in the task of salient object detection. 
 
\begin{figure}[H]
    \centering
    \includegraphics[width=1\linewidth]{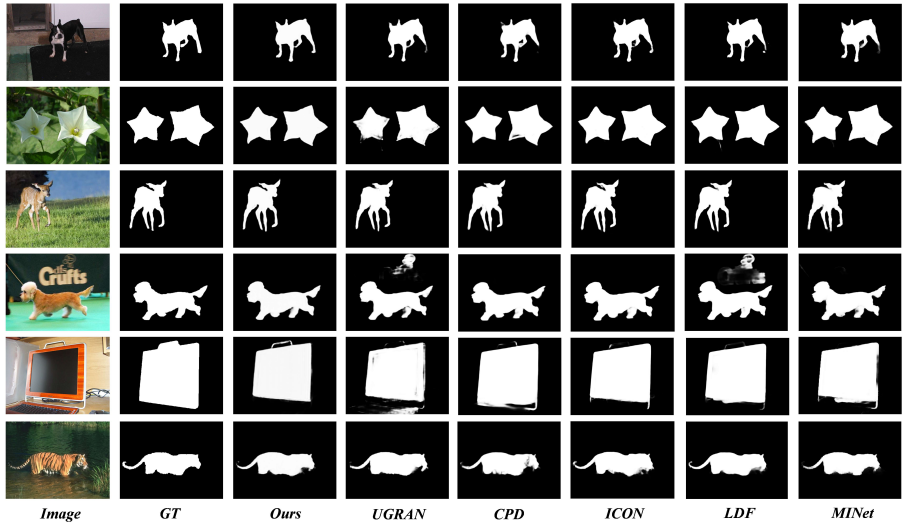}
    \caption{Visualization of comparative results for various saliency detection models. Our method better preserves salient structures and suppresses noise from shadows and low-saturation regions.}
    \label{fig:enter-label2}
\end{figure}

Figure~\ref{fig:enter-label2} provides a visual comparison of our model with other methods. It can be seen that our method has higher accuracy in restoring saliency maps and effectively reduces the interference of shadows and low-saturation areas. In addition, it shows better integrity and robustness when there are complex backgrounds or detailed structure maps.

\subsection{Ablation Study}

In order to further verify the effectiveness of each module, we use the basic model as a comparison, and gradually introduce DUGC and MCF modules to observe the effect of each module on the overall performance improvement. The experimental results are shown in Table~\ref{tab:ablation}, which shows that the introduction of each module has a positive effect on the performance improvement, verifying the effectiveness and necessity of the proposed design.

\renewcommand{\arraystretch}{1.2}

\begin{table}[]
\caption{Ablation study results across different datasets. Bold indicates the best result.}
\centering
\resizebox{\textwidth}{!}{%
\begin{tabular}{ccc|cccc|cccc|cccc}
\hline
\multicolumn{3}{c|}{Datasets} & \multicolumn{4}{c|}{ECSSD} & \multicolumn{4}{c|}{HKU-IS} & \multicolumn{4}{c}{SOD} \\ \hline
Base & DUGC & MCF & \textit{MAE}~\textcolor{red}{$\downarrow$} & \textit{mS~\textcolor{green}{$\uparrow$}} & $S_m$~\textcolor{green}{$\uparrow$} & $F_\beta^w$~\textcolor{green}{$\uparrow$} & \textit{MAE}~\textcolor{red}{$\downarrow$} & \textit{mS~\textcolor{green}{$\uparrow$}} & $S_m$~\textcolor{green}{$\uparrow$} & $F_\beta^w$~\textcolor{green}{$\uparrow$} & \textit{MAE}~\textcolor{red}{$\downarrow$} & \textit{mS~\textcolor{green}{$\uparrow$}} & $S_m$~\textcolor{green}{$\uparrow$} & $F_\beta^w$~\textcolor{green}{$\uparrow$} \\ \hline
\ding{51} & \ding{55} & \ding{55} & 0.034 & 0.856 & 0.924 & 0.911 & 0.028 & 0.856 & 0.922 & 0.905 & 0.094 & 0.665 & 0.795 & 0.758 \\
\ding{55} & \ding{51} & \ding{55} & \textbf{0.028} & 0.858 & 0.922 & 0.905 & 0.028 & 0.858 & 0.922 & 0.905 & 0.090 & 0.677 & 0.801 & 0.765 \\
\ding{55} & \ding{55} & \ding{51} & 0.031 & 0.872 & 0.930 & 0.915 & \textbf{0.026} & 0.873 & 0.928 & 0.912 & 0.082 & 0.697 & 0.818 & 0.780 \\ \hline
\ding{55} & \ding{51} & \ding{51} & 0.030 & \textbf{0.881} & \textbf{0.935} & \textbf{0.922} & 0.027 & \textbf{0.876} & \textbf{0.929} & \textbf{0.913} & \textbf{0.078} & \textbf{0.711} & \textbf{0.827} & \textbf{0.796} \\ \hline
\end{tabular}%
}
\label{tab:ablation}
\end{table}

Specifically, when the DUGC module is introduced alone, most of the indicators of the model on the three datasets are improved, indicating that the dynamic modeling uncertainty graph helps capture the complex relationships in the data. Similarly, the MCF module is introduced alone to bring performance improvements, which proves the effectiveness of multimodal collaborative fusion in enhancing feature representation and robustness.

More importantly, combining the DUGC and MCF modules achieves the best results on almost all indicators and datasets, indicating the complementary effects between the two components. This synergy enables the model to better utilize uncertainty propagation and multimodal information, resulting in more accurate and stable predictions.

\section{Limitation}
Although the proposed DUP-MCRNet achieves strong performance on multimodal saliency detection benchmarks, there remain limitations worth further exploration. First, the DUGC module introduces graph construction and multi-step uncertainty propagation, which, despite enhancing feature representation, incur high computational and memory overhead—especially with high-resolution inputs or multi-frame sequences—hindering real-time use on resource-limited platforms. Second, cross-domain generalization has yet to be fully validated. Existing experiments focus on standard datasets, but performance under real-world conditions—such as nighttime infrared or adverse weather—remains unclear, calling for further robustness evaluation.

\section{Conclusion}
This paper introduces DUP-MCRNet, a novel framework for salient object detection that tackles detail loss, edge ambiguity, and suboptimal multimodal fusion in complex visual scenes. The proposed Dynamic Uncertainty Graph Convolution (DUGC) module explicitly models uncertainty propagation via sparse spatial-semantic graphs, enabling adaptive refinement of small-scale structures and ambiguous boundaries. In parallel, the Multimodal Collaborative Fusion (MCF) strategy leverages learnable gating weights to dynamically integrate RGB, depth, and edge features, facilitating coherent global-local reasoning across modalities. Extensive experiments on five public benchmarks demonstrate that DUP-MCRNet consistently outperforms most of the state-of-the-art methods, especially in preserving edge sharpness and maintaining robustness in cluttered backgrounds. Ablation studies further validate the complementary strengths of the DUGC and MCF modules, showing their synergy significantly enhances boundary preservation and adaptability to challenging scenes. In future work, we aim to explore lightweight model deployment and extend the proposed uncertainty-aware fusion mechanisms to video-based saliency detection tasks.

\bibliography{ref}

\begin{thebibliography}{10}
\providecommand{\url}[1]{\texttt{#1}}
\providecommand{\urlprefix}{URL }
\providecommand{\doi}[1]{https://doi.org/#1}

\bibitem{cen2023sadsegmentrgbd}
Cen, J., Wu, Y., Wang, K., Li, X., Yang, J., Pei, Y., Kong, L., Liu, Z., Chen, Q.: Sad: Segment any rgbd (2023), \url{https://arxiv.org/abs/2305.14207}

\bibitem{8578420}
Chen, H., Li, Y.: Progressively complementarity-aware fusion network for rgb-d salient object detection. In: 2018 IEEE/CVF Conference on Computer Vision and Pattern Recognition. pp. 3051--3060 (2018). \doi{10.1109/CVPR.2018.00322}

\bibitem{chen2019reverseattentionsalientobject}
Chen, S., Tan, X., Wang, B., Hu, X.: Reverse attention for salient object detection (2019), \url{https://arxiv.org/abs/1807.09940}

\bibitem{6871397}
Cheng, M.M., Mitra, N.J., Huang, X., Torr, P.H.S., Hu, S.M.: Global contrast based salient region detection. IEEE Transactions on Pattern Analysis and Machine Intelligence  \textbf{37}(3),  569--582 (2015). \doi{10.1109/TPAMI.2014.2345401}

\bibitem{8954094}
Feng, M., Lu, H., Ding, E.: Attentive feedback network for boundary-aware salient object detection. In: 2019 IEEE/CVF Conference on Computer Vision and Pattern Recognition (CVPR). pp. 1623--1632 (2019). \doi{10.1109/CVPR.2019.00172}

\bibitem{Hou_2019}
Hou, Q., Cheng, M.M., Hu, X., Borji, A., Tu, Z., Torr, P.H.S.: Deeply supervised salient object detection with short connections. IEEE Transactions on Pattern Analysis and Machine Intelligence  \textbf{41}(4),  815–828 (Apr 2019). \doi{10.1109/tpami.2018.2815688}, \url{http://dx.doi.org/10.1109/TPAMI.2018.2815688}

\bibitem{10.1007/978-3-030-58523-5_4}
Ji, W., Li, J., Zhang, M., Piao, Y., Lu, H.: Accurate rgb-d salient object detection via collaborative learning. In: Vedaldi, A., Bischof, H., Brox, T., Frahm, J.M. (eds.) Computer Vision -- ECCV 2020. pp. 52--69. Springer International Publishing, Cham (2020)

\bibitem{b50}
Kingma, D.P., Ba, J.: Adam: A method for stochastic optimization (2017), \url{https://arxiv.org/abs/1412.6980}

\bibitem{b47}
Li, G., Yu, Y.: Visual saliency based on multiscale deep features. In: 2015 IEEE Conference on Computer Vision and Pattern Recognition (CVPR). pp. 5455--5463 (2015). \doi{10.1109/CVPR.2015.7299184}

\bibitem{b44}
Li, Y., Hou, X., Koch, C., Rehg, J.M., Yuille, A.L.: The secrets of salient object segmentation (2014), \url{https://arxiv.org/abs/1406.2807}

\bibitem{b52}
Liu, J.J., Hou, Q., Cheng, M.M., Feng, J., Jiang, J.: A simple pooling-based design for real-time salient object detection. In: 2019 IEEE/CVF Conference on Computer Vision and Pattern Recognition (CVPR). pp. 3912--3921 (2019). \doi{10.1109/CVPR.2019.00404}

\bibitem{9076883}
Liu, N., Han, J., Yang, M.H.: Picanet: Pixel-wise contextual attention learning for accurate saliency detection. IEEE Transactions on Image Processing  \textbf{29},  6438--6451 (2020). \doi{10.1109/TIP.2020.2988568}

\bibitem{9156287}
Liu, N., Zhang, N., Han, J.: Learning selective self-mutual attention for rgb-d saliency detection. In: 2020 IEEE/CVF Conference on Computer Vision and Pattern Recognition (CVPR). pp. 13753--13762 (2020). \doi{10.1109/CVPR42600.2020.01377}

\bibitem{liu2021visualsaliencytransformer}
Liu, N., Zhang, N., Wan, K., Shao, L., Han, J.: Visual saliency transformer (2021), \url{https://arxiv.org/abs/2104.12099}

\bibitem{9710580}
Liu, Z., Lin, Y., Cao, Y., Hu, H., Wei, Y., Zhang, Z., Lin, S., Guo, B.: Swin transformer: Hierarchical vision transformer using shifted windows. In: 2021 IEEE/CVF International Conference on Computer Vision (ICCV). pp. 9992--10002 (2021). \doi{10.1109/ICCV48922.2021.00986}

\bibitem{10697198}
Mao, Y., Zhang, J., Wan, Z., Tian, X., Li, A., Lv, Y., Dai, Y.: Generative transformer for accurate and reliable salient object detection. IEEE Transactions on Circuits and Systems for Video Technology  \textbf{35}(2),  1041--1054 (2025). \doi{10.1109/TCSVT.2024.3469286}

\bibitem{b40}
Movahedi, V., Elder, J.H.: Design and perceptual validation of performance measures for salient object segmentation. In: 2010 IEEE Computer Society Conference on Computer Vision and Pattern Recognition - Workshops. pp. 49--56 (2010). \doi{10.1109/CVPRW.2010.5543739}

\bibitem{pang2020multiscaleinteractivenetworksalient}
Pang, Y., Zhao, X., Zhang, L., Lu, H.: Multi-scale interactive network for salient object detection (2020), \url{https://arxiv.org/abs/2007.09062}

\bibitem{9010728}
Piao, Y., Ji, W., Li, J., Zhang, M., Lu, H.: Depth-induced multi-scale recurrent attention network for saliency detection. In: 2019 IEEE/CVF International Conference on Computer Vision (ICCV). pp. 7253--7262 (2019). \doi{10.1109/ICCV.2019.00735}

\bibitem{8953756}
Qin, X., Zhang, Z., Huang, C., Gao, C., Dehghan, M., Jagersand, M.: Basnet: Boundary-aware salient object detection. In: 2019 IEEE/CVF Conference on Computer Vision and Pattern Recognition (CVPR). pp. 7471--7481 (2019). \doi{10.1109/CVPR.2019.00766}

\bibitem{Qu_2017}
Qu, L., He, S., Zhang, J., Tian, J., Tang, Y., Yang, Q.: Rgbd salient object detection via deep fusion. IEEE Transactions on Image Processing  \textbf{26}(5),  2274–2285 (May 2017). \doi{10.1109/tip.2017.2682981}, \url{http://dx.doi.org/10.1109/TIP.2017.2682981}

\bibitem{sun2020miningcrossimagesemanticsweakly}
Sun, G., Wang, W., Dai, J., Gool, L.V.: Mining cross-image semantics for weakly supervised semantic segmentation (2020), \url{https://arxiv.org/abs/2007.01947}

\bibitem{b48}
Wang, L., Lu, H., Wang, Y., Feng, M., Wang, D., Yin, B., Ruan, X.: Learning to detect salient objects with image-level supervision. In: 2017 IEEE Conference on Computer Vision and Pattern Recognition (CVPR). pp. 3796--3805 (2017). \doi{10.1109/CVPR.2017.404}

\bibitem{wang2021salientobjectdetectiondeep}
Wang, W., Lai, Q., Fu, H., Shen, J., Ling, H., Yang, R.: Salient object detection in the deep learning era: An in-depth survey (2021), \url{https://arxiv.org/abs/1904.09146}

\bibitem{wang2021pyramidvisiontransformerversatile}
Wang, W., Xie, E., Li, X., Fan, D.P., Song, K., Liang, D., Lu, T., Luo, P., Shao, L.: Pyramid vision transformer: A versatile backbone for dense prediction without convolutions (2021), \url{https://arxiv.org/abs/2102.12122}

\bibitem{wei2019f3netfusionfeedbackfocus}
Wei, J., Wang, S., Huang, Q.: F3net: Fusion, feedback and focus for salient object detection (2019), \url{https://arxiv.org/abs/1911.11445}

\bibitem{b53}
Wei, J., Wang, S., Wu, Z., Su, C., Huang, Q., Tian, Q.: Label decoupling framework for salient object detection (2020), \url{https://arxiv.org/abs/2008.11048}

\bibitem{b51}
Wu, Z., Su, L., Huang, Q.: Cascaded partial decoder for fast and accurate salient object detection (2019), \url{https://arxiv.org/abs/1904.08739}

\bibitem{xiong2025nonstationarytimeseriesforecasting}
Xiong, Y., Wen, Y.: Non-stationary time series forecasting based on fourier analysis and cross attention mechanism (2025), \url{https://arxiv.org/abs/2505.06917}

\bibitem{b42}
Yan, Q., Xu, L., Shi, J., Jia, J.: Hierarchical saliency detection. In: 2013 IEEE Conference on Computer Vision and Pattern Recognition. pp. 1155--1162 (2013). \doi{10.1109/CVPR.2013.153}

\bibitem{b46}
Yang, C., Zhang, L., Lu, H., Ruan, X., Yang, M.H.: Saliency detection via graph-based manifold ranking. In: 2013 IEEE Conference on Computer Vision and Pattern Recognition. pp. 3166--3173 (2013). \doi{10.1109/CVPR.2013.407}

\bibitem{10.1007/978-981-96-2064-7_14}
Yu, J., Liu, Y., Wu, X., Xu, K., Li, J.: Pa2net: Pyramid attention aggregation network for saliency detection. In: Ide, I., Kompatsiaris, I., Xu, C., Yanai, K., Chu, W.T., Nitta, N., Riegler, M., Yamasaki, T. (eds.) MultiMedia Modeling. pp. 186--200. Springer Nature Singapore, Singapore (2025)

\bibitem{b54}
Yuan, Y., Gao, P., Dai, Q., Qin, J., Xiang, W.: Uncertainty-guided refinement for fine-grained salient object detection. IEEE Transactions on Image Processing  \textbf{34},  2301--2314 (2025). \doi{10.1109/TIP.2025.3557562}

\bibitem{Zhai_2021}
Zhai, Y., Fan, D.P., Yang, J., Borji, A., Shao, L., Han, J., Wang, L.: Bifurcated backbone strategy for rgb-d salient object detection. IEEE Transactions on Image Processing  \textbf{30},  8727–8742 (2021). \doi{10.1109/tip.2021.3116793}, \url{http://dx.doi.org/10.1109/TIP.2021.3116793}

\bibitem{8237293}
Zhang, P., Wang, D., Lu, H., Wang, H., Ruan, X.: Amulet: Aggregating multi-level convolutional features for salient object detection. In: 2017 IEEE International Conference on Computer Vision (ICCV). pp. 202--211 (2017). \doi{10.1109/ICCV.2017.31}

\bibitem{9008371}
Zhao, J., Liu, J.J., Fan, D.P., Cao, Y., Yang, J., Cheng, M.M.: Egnet: Edge guidance network for salient object detection. In: 2019 IEEE/CVF International Conference on Computer Vision (ICCV). pp. 8778--8787 (2019). \doi{10.1109/ICCV.2019.00887}

\bibitem{Zhou_2021}
Zhou, T., Fan, D.P., Cheng, M.M., Shen, J., Shao, L.: Rgb-d salient object detection: A survey. Computational Visual Media  \textbf{7}(1),  37–69 (Mar 2021). \doi{10.1007/s41095-020-0199-z}, \url{http://dx.doi.org/10.1007/s41095-020-0199-z}

\bibitem{10377606}
Zhou, Y., Li, Z., Guo, C.L., Bai, S., Cheng, M.M., Hou, Q.: Srformer: Permuted self-attention for single image super-resolution. In: 2023 IEEE/CVF International Conference on Computer Vision (ICCV). pp. 12734--12745 (2023). \doi{10.1109/ICCV51070.2023.01174}

\end{thebibliography}

\end{document}